\documentclass[runningheads]{llncs}
\usepackage[T1]{fontenc}
\usepackage{graphicx}
\usepackage{amsmath}
\usepackage{amssymb}
\usepackage{pifont}
\usepackage{booktabs}
\usepackage{subfigure}
\usepackage[dvipsnames]{xcolor}
\usepackage{hyperref}
\hypersetup{colorlinks=true,
citecolor=black,
urlcolor=magenta,
linkcolor=black,}

\begin{document}
%
\title{SHARP: Segmentation of Hands and Arms by Range using Pseudo-Depth for Enhanced Egocentric 3D Hand Pose Estimation and Action Recognition}
%
%
\author{Wiktor Mucha\inst{1}\orcidID{0000-0002-6048-3425} \and
Michael Wray\inst{2}\orcidID{0000-0001-5918-9029} \and
Martin Kampel\inst{1}\orcidID{0000-0002-5217-2854}}
\authorrunning{W. Mucha et al.}
\titlerunning{SHARP: Egocentric 3D Hand Pose and Action Recognition}
%


\institute{Computer Vision Lab, TU Wien, Favoritenstr. 9/193-1, 1040 Vienna, Austria \email{\{wiktor.mucha,martin.kampel\}@tuwien.ac.at}\and
University of Bristol \\
\email{michael.wray@bristol.ac.uk}\\ 
}
\maketitle              
\begin{abstract}

Hand pose represents key information for action recognition in the egocentric perspective, where the user is interacting with objects. We propose to improve egocentric 3D hand pose estimation based on RGB frames only by using pseudo-depth images. Incorporating state-of-the-art single RGB image depth estimation techniques, we generate pseudo-depth representations of the frames and use distance knowledge to segment irrelevant parts of the scene. The resulting depth maps are then used as segmentation masks for the RGB frames. Experimental results on \textit{H2O Dataset} confirm the high accuracy of the estimated pose with our method in an action recognition task.
The 3D hand pose, together with information from object detection, is processed by a transformer-based action recognition network, resulting in an accuracy of 91.73\%, outperforming all state-of-the-art methods. Estimations of 3D hand pose result in competitive performance with existing methods with a mean pose error of 28.66 mm. This method opens up new possibilities for employing distance information in egocentric 3D hand pose estimation without relying on depth sensors. The code is available under \url{https://github.com/wiktormucha/SHARP}

\end{abstract}
\section{Introduction}
\label{sec:introduction}
\begin{figure}[t]
\includegraphics[width=\textwidth]{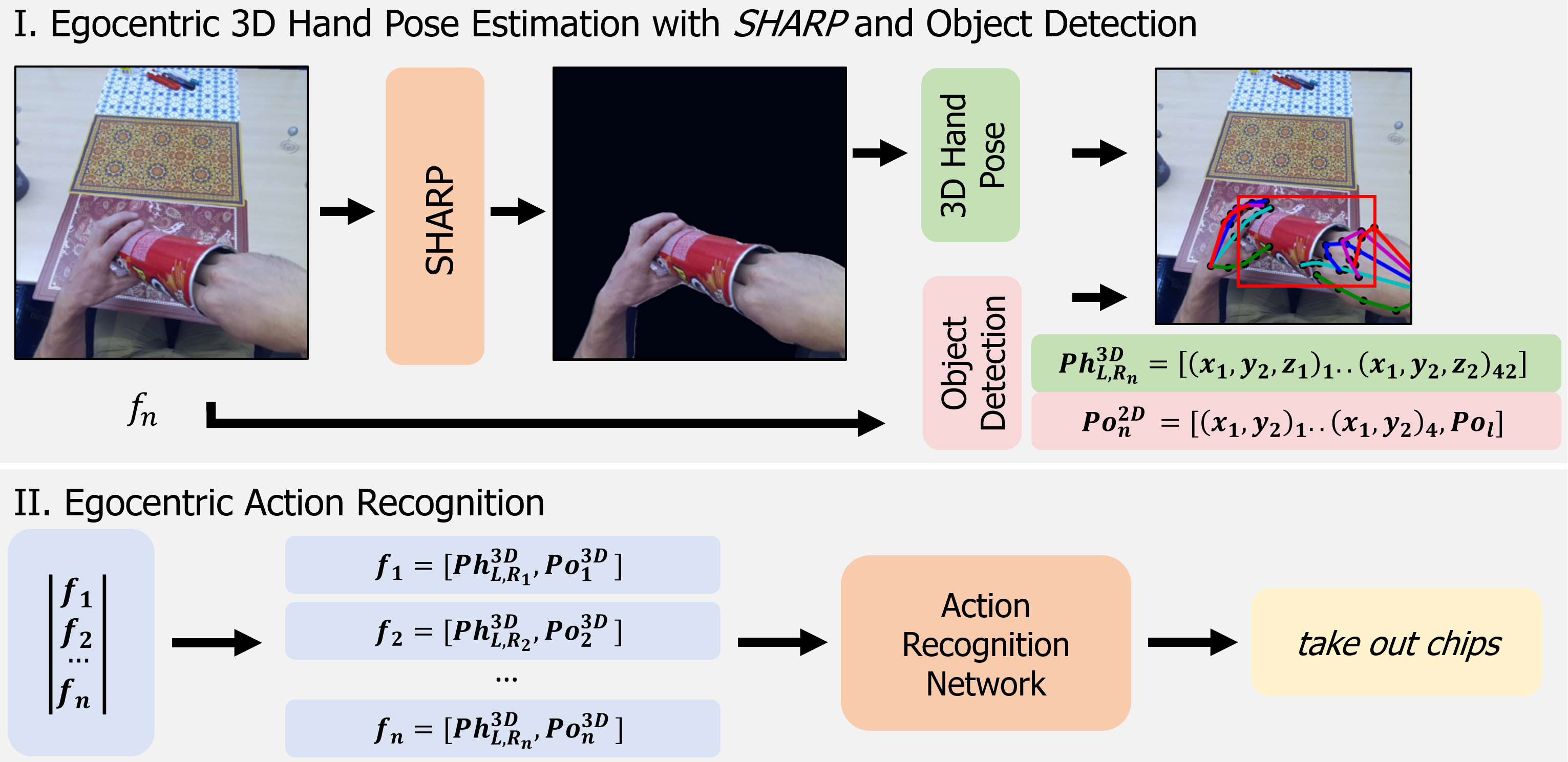}
\caption{Overview of our method. In the sequence of input frames $f_1, f_2, f_3\dots f_n$ representing the action, \textit{SHARP} improves the estimation of the 3D hand pose $Ph^{3D}_{L,R,n}$. The bounding box of the manipulated objects $Po^{2D}_{n}$ with their labels $Po_{l}$ are retrieved using \textit{YOLOv7} \cite{wang2022yolov7}. Pose information is embedded in a vector describing each frame. The sequence of vectors is processed by the transformer-based network to predict action.}
\label{fig:teaser}
\end{figure}

In recent years, one of the growing research areas in computer vision has been egocentric vision, as evidenced by the increasing number and size of published datasets \textit{EPIC-KITCHENS}~\cite{Damen2018EPICKITCHENS}, \textit{Ego4D}~\cite{grauman2022ego4d}, \textit{H2O}~\cite{Kwon_2021_ICCV} and release of devices like Ray-Ban Stories, Apple Vision Pro or HoloLens. 
One of the challenges in egocentric vision is understanding human-object interaction based on hand pose estimation and action recognition~\cite{garcia2018first,Kwon_2021_ICCV}.
The hand pose estimation task is described as the challenge of estimating the position of key points representing the joints of a human hand in two or three-dimensional space. Estimated positions are a valuable source of information for recognising the actions performed by a camera wearer, linking these two tasks. 
Egocentric action recognition research is of great importance in various domains, including augmented and virtual reality, nutritional behaviour analysis, and Active Assisted Living (AAL) technologies for lifestyle analysis~\cite{nunez2022egocentric} or assistance~\cite{text2taste}. As AAL technologies mainly target Activities of Daily Living (ADLs) such as drinking, eating and food preparation, which are inherently manual and involve object manipulation, there's a growing interest in research focused on hand-based action recognition.

Current work on egocentric hand-based action recognition focuses on 3D hand pose~\cite{tekin2019h+,das2021symmetric,Kwon_2021_ICCV} using a single RGB camera. As a result, these studies regress $z$ coordinate from RGB frames, which introduces complexity and results in pose prediction errors of around 40 mm~\cite{tekin2019h+,hasson2020leveraging,Kwon_2021_ICCV} (equivalent to a 20.5\% error considering an average human hand size of 18 cm), which is far from the desired performance, especially considering that publicly available datasets for egocentric hand pose are captured in a laboratory environment. Accurate pose prediction is essential for hand-based action recognition~\cite{effhandegonet}. The improvement in 3D prediction could be further enhanced by the use of a depth sensor, but there's currently no portable depth sensor on the market. Despite market availability, an additional sensor would add undesired costs due to power and processing requirements. Data growth for training and research is another constraint, as labelling key points in 3D space is difficult and requires, for example, a laboratory multi-view camera setup~\cite{Kwon_2021_ICCV,ohkawa:cvpr23}. All these circumstances create a need and motivate our research to explore new techniques and solutions to improve egocentric 3D pose estimation based on RGB images only.

Our study proposes the use of pseudo-depth images, depth images generated from a single RGB image using state-of-the-art depth estimation methods. The resulting distance representation of the scene does not contain real depth values, but it allows for the removal of non-relevant information in the scene depending on the distance. In an egocentric perspective, human arms have a constant maximum distance from the camera because the camera is mounted in a fixed position on the human body. This characteristic allows for the removal of the values representing the parts of the scene beyond this distance, leaving the input image of a hand pose estimation network with only hands and manipulated objects visible. We call this process Segmentation of Hands and Arms by Range using Pseudo-depth (SHARP). This solution requires no additional sensors; it can be applied to any RGB input data; no additional training of the depth estimation model is required; and compared to background subtraction based on image sequences, only a single RGB image is required. These advantages are confirmed by a performance improvement of 7 mm, reducing the mean pose error from 35.48 mm to 28.66 mm from the baseline. The overview of the method is presented in Fig. \ref{fig:teaser}. Our contribution can be listed as follows:
\begin{itemize}
    \item Inspired by superior egocentric hand pose estimation in 2D over other methods, we extend the state-of-the-art \textit{EffHandEgoNet} \cite{effhandegonet} to 3D pose estimation, resulting in a new architecture called \textit{EffHandEgoNet3D}.
    
    \item 
    
    On the top of \textit{EffHandEgoNet3D} we propose \textit{SHARP module}, a novel idea for egocentric scene segmentation to improve hand-object interaction understanding. A state-of-the-art depth estimation model is used to generate a pseudo-depth scene representation. Furthermore, the generated distance knowledge is used to remove irrelevant information in the scene with a fixed distance over the range of the human arms, resulting in the preservation of the human arms and the interacting object. \textit{SHARP} requires no additional training and can be applied to any egocentric RGB data. The proposed architecture outperforms several state-of-the-art studies, achieving a mean error of 28.66 mm on the \textit{H2O Dataset}.

    \item We implement an action recognition network based on a transformer architecture. It uses previously estimated 3D hand pose and 2D object detection information as input. The network outperforms the state-of-the-art on the \textit{H2O Dataset}, including methods that use more information e.g. 6D object pose, reaching 91.73\% action recognition accuracy.
    
    \item We present extensive experiments and ablations performed on \textit{H2O Dataset}, showing the influence of the proposed scene segmentation method on the performance of 3D hand pose estimation in the egocentric perspective.
\end{itemize}

The structure of the paper is as follows: In section \ref{sec:related_work}, we review related research on egocentric 3D hand keypoint estimation, hand-based action recognition, and depth estimation using a single RGB image, and identify opportunities for improvement. Section \ref{sec:methodology} details our approach and its implementation. Our evaluation and experimental results are presented in section \ref{sec:evaluation}. Finally, section \ref{sec:conclusion} concludes the study, summarising its main findings and limitations.

\section{Related Work}
\label{sec:related_work}
\paragraph{\textbf{Egocentric Hand Pose Estimation}}

Hand pose estimation in egocentric vision faces challenges such as self-occlusion, limited field of view, and diverse perspectives, which hinder effective generalisation. Some approaches overcome these obstacles by using RGB-D sensors \cite{mueller2017real,yamazaki2017hand,garcia2018first}. However, the adoption of depth sensors is hampered by limited market availability, directing towards self-made solutions and increasing computing and power costs. Due to device limitations, several studies estimate 3D keypoints from RGB images only by using neural networks that estimate the z coordinate representing depth along x and y, followed by a conversion from 2D to 3D space using intrinsic camera parameters \cite{tekin2019h+,Kwon_2021_ICCV}. 
For example, Tekin et al. \cite{tekin2019h+} compute the 3D pose of a hand directly from a single RGB image using a convolutional neural network (CNN) that outputs a 3D grid with the probability of target pose values in each cell. Similarly, Kwon et al. \cite{Kwon_2021_ICCV} extend this approach to estimate poses for both hands. However, these methods report a mean end-point error (EPE) of 37 mm for hand pose estimation in the \textit{H2O dataset}, suggesting room for improvement given the average human hand size of 18 cm. Cho et al. \cite{cho2023transformer} use CNNs with transformer-based networks for 3D pose reconstruction on a frame-by-frame basis, while Wen et al. \cite{wen2023hierarchical} propose a sequence-based approach for depth reconstruction that addresses occlusion challenges.

\paragraph{\textbf{Egocentric Action Recognition}}

A common strategy for action recognition involves the joint processing of hand and object information. Cartas~et~al.~\cite{cartas2017contextually} proposes CNN-based object detectors to estimate the positions of primary regions (hands) and secondary regions (objects). Temporal information from these regions is then processed by a Long Short-Term Memory (LSTM) network. Nguyen~et~al.~\cite{nguyen2019neural} Transition from bounding box information to 2D skeletons of a single hand estimated by CNN from RGB and depth images. The joints of these skeletons are aggregated using spatial and temporal Gaussian aggregation, and action recognition is performed using a learnable Symmetric Positive Definite (SPD) matrix. With the rise of 3D-based hand pose estimation algorithms, the scientific community has increasingly focused on egocentric action understanding using 3D information~\cite{tekin2019h+,das2021symmetric,Kwon_2021_ICCV}. Tekin~et~al.~\cite{tekin2019h+} estimate 3D hand and object poses from a single RGB frame using a CNN, embedding temporal information to predict action classes using an LSTM. Other techniques use graph networks, such as Das~et~al.~\cite{das2021symmetric}, who present a spatio-temporal graph CNN architecture that describes finger motion using separate subgraphs. Kwon~et~al.~\cite{Kwon_2021_ICCV} construct sub-graphs for each hand and object, which are merged into a multigraph model, allowing learning of interactions between these components. Wen~et~al.~\cite{wen2023hierarchical} use a transformer-based model with estimated 3D hand pose and object label input. Cho~et~al.~\cite{cho2023transformer} enrich the transformer inputs with object pose and hand-object contact information. However, these studies do not make use of depth data. Instead, they estimate points in 3D space using neural networks and intrinsic camera parameters~\cite{tekin2019h+,Kwon_2021_ICCV,wen2023hierarchical,cho2023transformer}.

\paragraph{\textbf{Depth Estimation from Single RGB Image}}

Recent advances in depth estimation have relied on CNNs for direct regression of scene depth from input images \cite{eigen2014depth}. These methods often struggle to generalise to unconstrained scenes due to the limited diversity and size of the training data. Garg et al. \cite{garg2016unsupervised} proposed the use of calibrated stereo cameras for self-supervision, which simplifies data acquisition but maintains constraints on specific data regimes. Despite subsequent self-supervised approaches \cite{godard2019digging}, challenges remain, particularly for dynamic scenes. Efforts to overcome these limitations include crowd-sourced annotation of ordinal relationships \cite{chen2016single}, but existing datasets are often biased or lack dynamic objects, making it difficult to generalise to less constrained environments. In response, Ranftl et al. \cite{ranftl2020towards} propose tools for mixing multiple datasets, even with incompatible annotations. Their approach incorporates a robust training objective, principled multi-objective learning, and emphasises pre-training of encoders on ancillary tasks. By training on five different sources, including a rich dataset of 3D movies, they outperform state-of-the-art depth estimation models in zero-shot cross-dataset performance.
As an extension of this work, Ranftl et al. \cite{ranftl2021vision} present \textit{DPT-Hybrid} and \textit{DPT-Large} architectures enhanced with dense prediction transformers, which use vision transformers instead of CNNs, further improving the performance of depth estimation.

\paragraph{\textbf{
What distinguishes our work}} from other studies of egocentric 3D hand pose is the use of a depth estimation that we incorporate into \textit{SHARP} module. Using state-of-the-art single RGB image depth estimation techniques, we generate a pseudo-depth representation of the image without any additional equipment. Knowing that the distance of the human arms from the camera in an egocentric view is constant, we then use this generated depth image to segment irrelevant information from the scene using a fixed distance threshold, thereby unifying the dataset for hand pose estimation. This methodology ensures that the hand pose estimation model only considers hands and manipulated objects, thereby increasing accuracy and efficiency, and can be applied to any RGB dataset.

\section{Egocentric 3D Hand Pose Estimation and Action Recognition Enforced With Pseudo Depth}
\label{sec:methodology}
\begin{figure}[t]
\includegraphics[width=\textwidth]{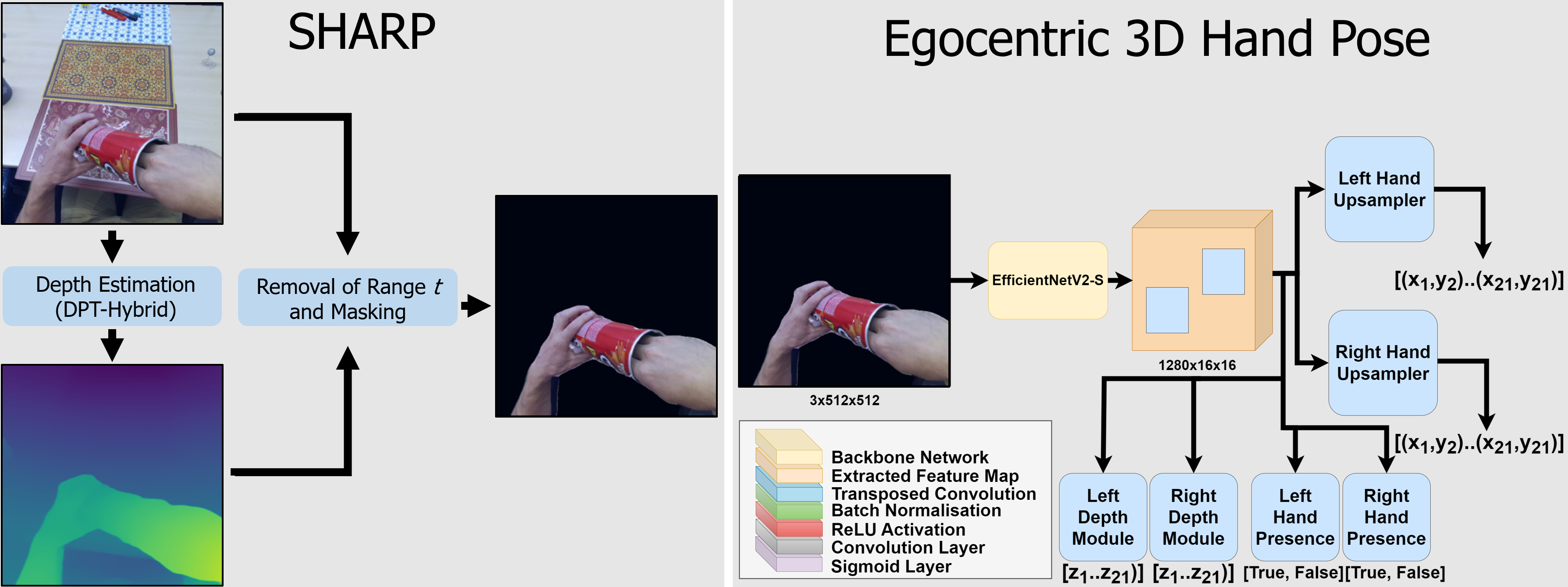}
\caption{
Overview of the proposed egocentric 3D hand pose estimation method. First, the RGB image is processed with the \textit{SHARP} module. Within \textit{SHARP}, the pseudo-depth image is generated using the \textit{DPT-Hybrid}. This distance representation is used to remove irrelevant scene information using a fixed threshold of the human arm range~$t$. Secondly, the \textit{SHARP} output is passed through a 3D hand pose estimation network.
} 
\label{fig:handpose_overview}
\end{figure}

The study considers the tasks of egocentric 3D hand pose estimation and action recognition. These two tasks are correlated but significantly different, so the methodology is described separately for each. 

\subsection{Egocentric 3D Hand Pose with Pseudo-Depth Segmentation}

In the first stage, each RGB frame $f_n$ undergoes processing with \textit{SHARP} module which consists of a depth estimation model \textit{DPT-Hybrid} \cite{ranftl2021vision}, yielding a pseudo-depth representation $I^D_n$ of the frame $f_n$. This pseudo-depth map is then normalised with its maximum value ${max(I^D_n)} $. As human arms have a constant maximum range we utilise this characteristic. Subsequently, a fixed threshold $t$ is applied to the pseudo depth map $I^D_n$ to remove the non-relevant scene part. The resultant depth map, devoid of background interference, serves as a segmentation mask for the $f_n$. Segmentation of $f_n$ with $I^D_n$ results in $I^{SEG}_n$ where the RGB image contains only human arms and a manipulated object.

The processed $ I^{SEG}_n \in \mathbb{R}^{3\times w\times h}, w,h = 512$ is then inputted into a 3D hand pose estimation network, named \textit{EffHandEgoNet3D}, which is an extension of the state-of-the-art 2D egocentric hand pose network, \textit{EffHandEgoNet} \cite{effhandegonet}, tailored for 3D estimation. \textit{EffHandEgoNet3D} comprises an \textit{EfficientNetV2-S} \cite{tan2021efficientnetv2} backbone which extract feature map representation of $I^{SEG}_n$ $F_M \in \mathbb{R}^{1280\times 16 \times 16}$. Extracted feature map $F_M$ is handed to two independent upsamplers for each of the hands and $MLP^{Z}_{L,R}$ estimating keypoints' depth. Despite pose estimation, the handness modules responsible for predicting each hand's presence $h_L, h_R \in \mathbb{R}^2$ are built from another $MLP^H_{L,R}$. The upsamplers consist of three transposed convolutions with batch normalisation and ReLU activation except the last layer followed by a pointwise convolution. Output results are heatmaps $\mathit H_{L,R}\in \mathbb{R}^{J\times w\times h}$ where each cell represents the probability of joint $J$ occurrence for each hand. In the next step they are transformed into $\mathit P^{2D}_{L,R}$ and concatenated with estimated corresponding $z$ values resulting in $\mathit P^{2.5D}_{L,R}$. The final step utilises camera intrinsic parameters to transform $\mathit P^{2.5D}_{L,R}$ using the pinhole camera model to camera space resulting in $\mathit P^{3D}_{L,R}$. The overview of the complete method is visible in Figure \ref{fig:handpose_overview}.

\begin{figure}[t]
\includegraphics[width=\textwidth]{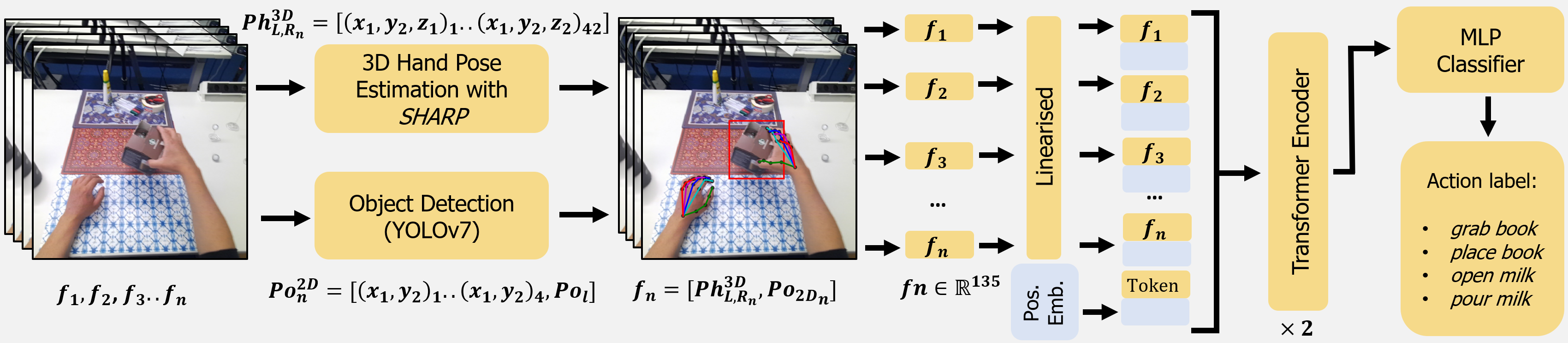}
\caption{
Our action recognition procedure. From the sequence of frames $f_1, f_2, f_3 ... f_n$ the hand pose $Ph^{3D}_{L,R}$ is estimated with \textit{SHARP} and \textit{EffHandEgoNet3D} model and the object pose $Po^{2D}$, $Po_{l}$ is extracted with \textit{YOLOv7} \cite{wang2022yolov7}. Each sequence frame $f_n$ is linearised and positional embedding and classification tokens are added. Next, this sequence is passed to a transformer encoder~\cite{dosovitskiy2020image} repeated $\times2$ times, which embeds the temporal information. Finally, the MLP predicts one of the 36 action labels.
}
\label{fig:ar_method}
\end{figure}

\subsection{Egocentric Action Recognition based on 3D Hand Pose}

We perform egocentric action recognition from image sequences using estimated 3D hand pose and 2D information about interacting object. The actions considered in this study are those in which humans manipulate objects with one or both hands, such as \textit{pouring milk} or \textit{opening a bottle}. An overview of the pipeline is shown in Fig. \ref{fig:ar_method}. It consists of three distinct components: object detection, 3D hand pose estimation, and finally action recognition using a transformer encoder and a classification MLP. The architecture improves egocentric action recognition based on the 2D hand pose introduced in \textit{EffHandEgoNet} study \cite{effhandegonet}. The first step in the pipeline is object detection, which is carried out employing the pre-trained \textit{YOLOv7} network \cite{wang2022yolov7}. In each frame, denoted as $\mathit f_n$, the interacting object is represented by $\mathit{Po}_{2D}(x,y) \in \mathbb{R}^{4 \times 2}$, where each point corresponds to the corners of its bounding box. Additionally, $\mathit{Po}_{l} \in \mathbb{R}^{1}$ represents object's label.

The representation of each action sequence consists of frames $\mathit [f_1,f_2,f_3,...,f_n]$, where $n \in [1..N]$ and $N=20$ following \cite{effhandegonet}. These frames embed flattened poses of hands $Ph^{3D}_{L,R}$ and object $Po_{2D}$,  $Po_{l}$. If fewer than $N$ frames represent an action, zero padding is applied, while actions longer than $N$ frames are sub-sampled.
The input vector $\mathit V_{seq}$ is a concatenation of frames $\mathit f_n \in \mathbb{R}^{135} $.
\begin{equation}
f_n =[ Ph^{3D}_{L}, Ph^{3D}_{R}, Po_{2D}, Po_{l}]
\end{equation}
\begin{equation}
V_{seq} = [f_1, f_2..f_n], \; n \in [1..N]
\end{equation}

The sequence vector representing an action $\mathit V_{seq}$ is processed to embed temporal information with a transformer encoder block following \cite{effhandegonet}. First, $\mathit V_{seq}$ is linearised using a fully connected layer to $\mathit x_{lin}$. The resulting $\mathit x_{lin}$ is combined with a classification token and a positional embedding.
The embedded sequence is passed to MLP for classifying the action.

\section{Experiments}
\label{sec:evaluation}
\subsection{Datasets}

In this evaluation, we focus exclusively on the \textit{H2O Dataset} \cite{Kwon_2021_ICCV} due to its suitability for our research objectives. This dataset captures human actions from an egocentric perspective, providing labels for action recognition and 3D hand pose of both hands. At the time of this study, there are only two other publicly available datasets with similar characteristics required for our study, such as \textit{AssemblyHands} \cite{ohkawa:cvpr23} and \textit{HoloAssist} \cite{HoloAssist2023}. While \textit{HoloAssist} is potentially valuable, the hand pose labels have not yet been released. \textit{AssemblyHands} is excluded due to images captured by infrared cameras, which are incompatible with the \textit{DPT-Hybrid} depth estimation model designed for RGB input. 

\paragraph{\textit{\textbf{H2O Dataset}}} is a comprehensive resource for analysing hand-based actions and object interactions involving two hands. It includes multi-view RGB-D images annotated with action labels covering 36 different classes derived from verb and object labels. It also includes 3D poses for both hands, resulting in $j = 2 \times 21$ points, and 6D poses and meshes for the manipulated objects. Ground truth camera poses and scene point clouds further enrich the dataset. The actions captured in the dataset were performed by four people. For both the action recognition and hand pose estimation tasks, the dataset provides training, validation and test subsets. The action recognition subset contains 569 clips for training, 122 for validation and 242 for testing.

\subsection{Metrics}

To evaluate the hand pose estimation and compare our work with the state of the art, we calculate the Mean Per Joint Position Error (MPJPE) in millimetres over 21 keypoints $J$ representing the human hand. This error metric quantifies the Euclidean distance between the predicted and ground truth values. For action recognition, we use the top-1 accuracy measure, where the model's prediction must exactly match the expected ground truth to be considered accurate.

\subsection{Experiment setup}

For both learning processes, each run is repeated three times to reduce the effect of random initialisation of the network, and mean results with standard deviations are reported.

\paragraph{\textbf{3D Hand Pose Estimation}} is trained and evaluated on \textit{H2O Dataset}. The optimisation is done using Stochastic Gradient Descent (SGD) over the summarised loss function including Intersection over Union (IoU) for each upsampler and $L1$ loss for predicted corresponding depth values. The process starts with a learning rate $l_r = 0.1$ and momentum equal to $m=0.9$. Over time $l_r$ is reduced by $\alpha = 0.5$ every $10^{th}$ epoch starting from the $50^{th}$ epoch. the data is augmented with random cropping, horizontal flipping, vertical flipping, resizing, rotating and blurring. The batch size is equal to $b_s = 32$. Model weights are saved for the smallest MPJPE in the validation subset.

\paragraph{\textbf{Action Recognition}}

module requires object detection. For this, we fine-tune YOLOv7 on the \textit{H2O Dataset} using the open-source strategy reported by the authors. The training of the action recognition includes the augmentation of the sequence vectors with keypoints using random rotation and an additional strategy with random masking of either the hand, the object positions or the label. This is done by setting the corresponding values of the hand or object in the frame $\mathit f_n$ to zero. We follow \cite{Kwon_2021_ICCV,tekin2019h+,yan2018spatial} and use given poses in training. Input sequence frames are randomly sub-sampled during training and uniformly sub-sampled for validation and testing. Models are trained with a batch size $b_s = 64$, AdamW optimiser, cross-entropy loss function, and a learning rate $l_r = 0.001$ reduced by a factor of 0.5 every 200 epochs after 500 epochs. Hyperparameters and augmentations are selected based on the best-performing set in the validation subset. Weights are stored for best validation accuracy.

\subsection{Comparison with State of the Art}

\begin{table}[b]
\centering

\caption{Results of 3D hand pose estimation provided in \textit{mm} in camera space.}
\label{tab:hand_pose}

\begin{tabular*}{\textwidth}{l@{
\extracolsep{\fill}}cccc}
\toprule

Method:       & Year    & MPJPE Left $\downarrow$& MPJPE Right $\downarrow$& MPJPE Both $\downarrow$ \\
\midrule
LPC \cite{hasson2020leveraging} &2020    & 39.56       & 41.87 &   40.72    \\ 

H+O \cite{tekin2019h+}  &2019   & 41.42       & 38.86 & 40.14      \\ 

H2O \cite{Kwon_2021_ICCV} &2021    & 41.45       & 37.21 & 39.33      \\ 

HTT \cite{wen2023hierarchical}  &2023   & 35.02       & 35.63 & 35.33      \\

H2OTR \cite{cho2023transformer} &2023 & 24.40 & 25.80 & \textbf{25.10} \\

THOR-Net \cite{aboukhadra2023thor-net} &2023 & 36.80 & 36.50 &36.65 \\

\textbf{Ours} &Now & 30.31     & 27.02  & 28.66    \\ 
\bottomrule
\end{tabular*}%

\end{table}

\begin{figure}[t]
\includegraphics[width=\textwidth]{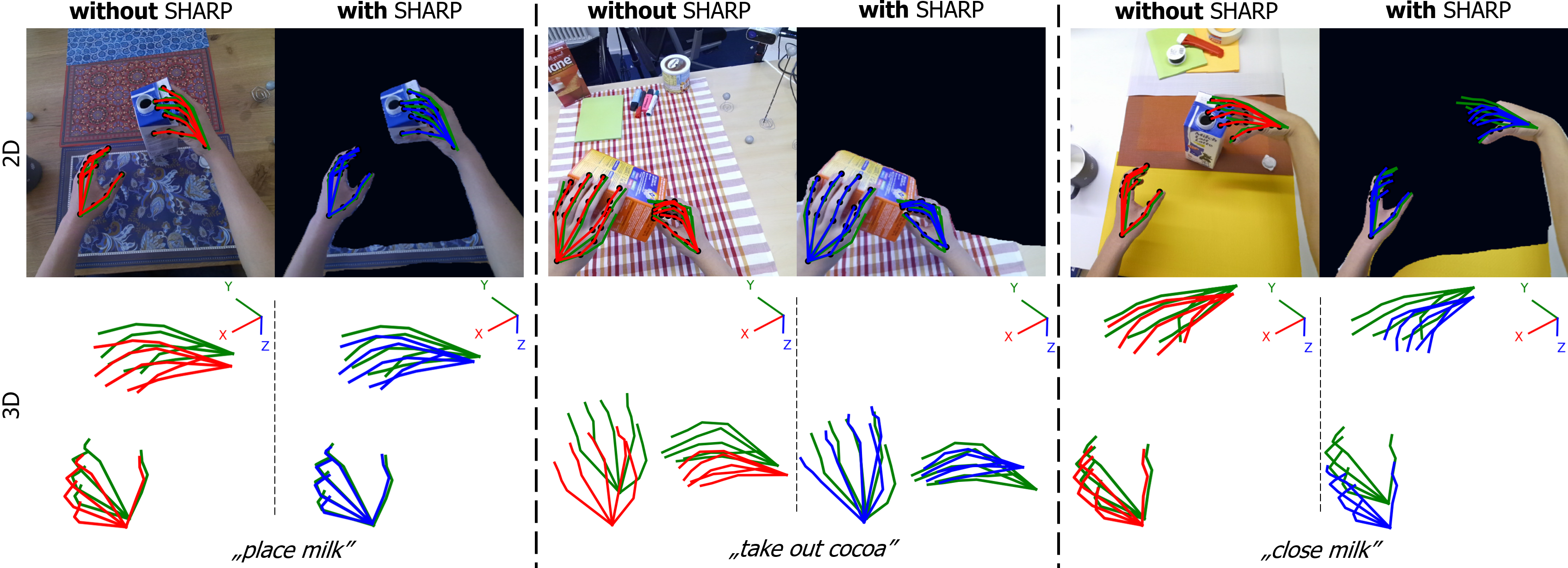}
\caption{
Qualitative results of our method in 2D and 3D space. Green skeletons represent the \textcolor{ForestGreen}{ground truth hand pose}, red estimations \textcolor{red}{\textbf{without} \textit{SHARP}} and blue estimations \textcolor{blue}{\textbf{with} \textit{SHARP}}. Images are annotated with a predicted action label for the represented sequences. Two examples from the left show that \textit{SHARP} improves 3D pose estimation. On the right, the 3D error increases as \textit{SHARP} partially loses the right hand. 
}
\label{fig:qualitative_results}
\end{figure}

\begin{figure}[t]
\includegraphics[width=\textwidth]{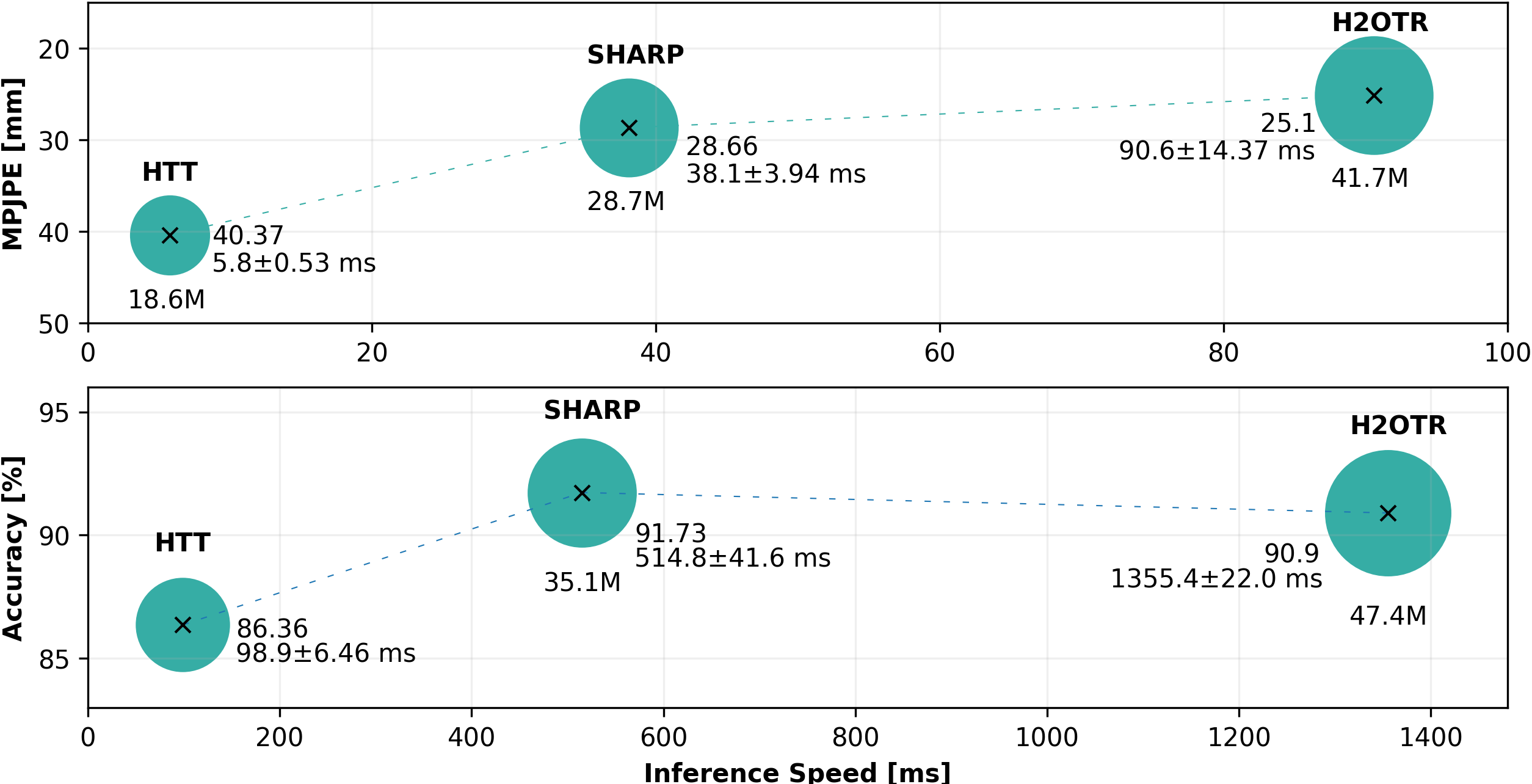}
\caption{Inference time for 3D hand pose estimation per single frame and action recognition accuracy per single action of state-of-the-art methods on \textit{H2O Dataset}. Each method is visualised as a circle whose size represents the number of trainable parameters. \textit{SHARP} inference is $\approx\times2.5$ faster than H2OTR \cite{cho2023transformer} with better action recognition.} 
\label{fig:inference}
\end{figure}

\begin{table}[t]
\centering
\caption{Results in accuracy of action recognition methods on \textit{H2O Dataset}. Inputs of methods are: \textit{Img} stands for semantic features extracted from an image using CNN network, \textit{Hand Pose} and \textit{Obj Ppose} stand for pose information type for hands and objects, and \textit{Obj Label} stands for object label. Results origin from referenced studies.}
\label{tab:results}
\begin{tabular*}{\textwidth}{l@{
\extracolsep{\fill}}cccccc}

\toprule
Method:   & Year & Img & Hand Pose & Obj Pose  & Obj Label  & Acc. $\uparrow$ \\
\midrule
C2D \cite{wang2018non} & 2018 & \checkmark     & \ding{55}  & \ding{55} & \ding{55}      & 70.66  \\
I3D \cite{carreira2017quo} & 2017 & \checkmark & \ding{55} &\ding{55}  & \ding{55}           & 75.21   \\
SlowFast \cite{feichtenhofer2019slowfast} & 2019 & \checkmark & \ding{55} &\ding{55}  & \ding{55}     & 77.69 \\
H+O \cite{tekin2019h+}  & 2019   &  \ding{55} & 3D  & 6D & \checkmark      & 68.88  \\
ST-GCN \cite{yan2018spatial} & 2018 &   \ding{55} & 3D & 6D & \checkmark        & 73.86\\
TA-GCN \cite{Kwon_2021_ICCV} & 2021 & \ding{55} & 3D &6D & \checkmark        & 79.25\\
HTT \cite{wen2023hierarchical}  & 2023 & \checkmark & 3D & \ding{55}   & \checkmark   & 86.36\\ 
H2OTR \cite{cho2023transformer} & 2023 & \ding{55} & 3D & 6D   & \checkmark   & 90.90\\ 
EffHandEgoNet \cite{effhandegonet} & 2024 & \ding{55}  & 2D & 2D & \checkmark & 91.32 \\
\textbf{Ours} & Now & \ding{55}  & 3D & 2D & \checkmark & \textbf{91.73} \\
\bottomrule
\end{tabular*}
\end{table}

Our architecture with \textit{SHARP} gives an average MPJPE in hand pose of $29.61\pm0.71$ mm in three consecutive runs with the best run MPJPE equal to 28.66 mm. The qualitative results shown in Fig. \ref{fig:qualitative_results} confirm the improvement in 3D hand pose estimation when using \textit{SHARP}, but also show that \textit{SHARP} can lead to a degradation in performance if too much information is reduced from the scene. Further, we employ the estimated 3D hand pose using \textit{SHARP} in the proposed action recognition architecture. It yields an average of 90.90\%$\pm$0.67 over three runs, with the best model yielding an accuracy of 91.73\%. 
Comparison with state of the art for egocentric 3D hand pose estimation is presented in Table \ref{tab:hand_pose}. 
Table \ref{tab:results} presents a comparison of state-of-the-art action recognition methods and their results on the \textit{H2O Dataset} reported by the authors. To ensure a fair comparison, the table provides details regarding the inputs of the action recognition modules.
For both tasks, we follow other studies~\cite{tekin2019h+,aboukhadra2023thor-net,wen2023hierarchical,cho2023transformer,effhandegonet} and report our best results.

We measure the inference times of our methods for the hand pose estimation task for a single frame and for a complete action recognition pipeline for a single action. The evaluation is performed by averaging the inference times over 1000 trials on the NVIDIA GeForce RTX3090 GPU for reliability. The results are shown in Fig. \ref{fig:inference}, where the upper part shows the hand pose performance and the lower part shows the action recognition. Our methods are compared with HTT \cite{wen2023hierarchical} and H2OTR \cite{cho2023transformer} as they are the only open-source implementations that allow such a comparison on the \textit{H2O Dataset} at the time of this study. 

\textit{SHARP} estimates the egocentric 3D hand pose with the second best result, being faster $\approx \times2.4$ than the best H2OTR~\cite{cho2023transformer} with 13M fewer parameters and only a 3mm performance penalty. Our action recognition outperforms all state-of-the-art methods and infers $\approx \times 2.6$ faster with 12M fewer parameters than the second best H2OTR~\cite{cho2023transformer}.

\subsection{Ablation Studies}

\begin{figure}[t]
\includegraphics[width=\textwidth]{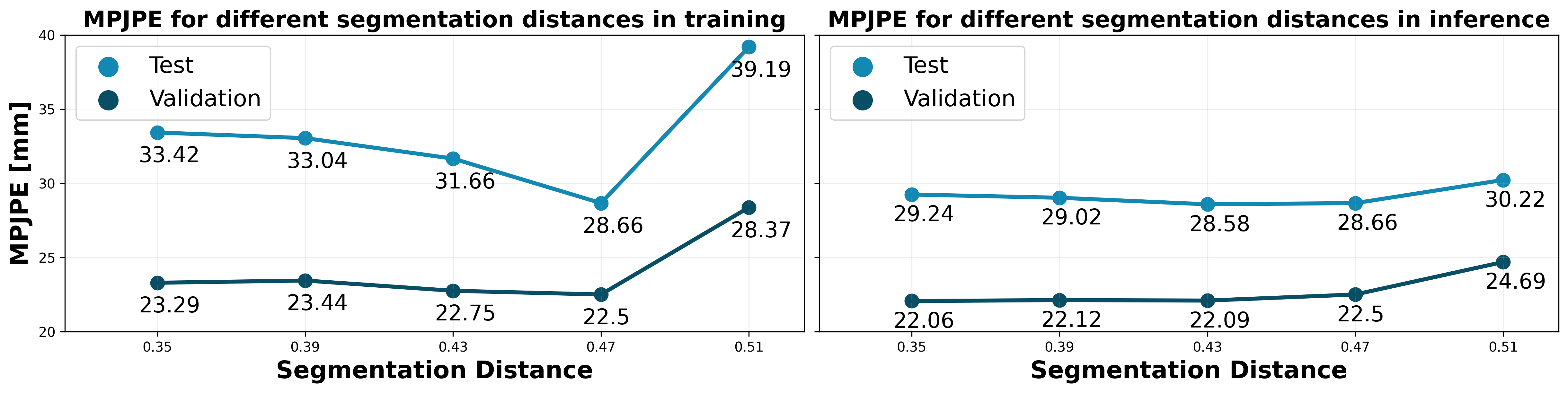}
\caption{Figures showing the results of the 3D hand pose estimation error in MPJPE as a function of the segmentation threshold $t$. The left figure shows the performance with different thresholds used for training and the right figure shows the performance for the best trained model with $t = 0.47$ and different $t$ during inference.} 
\label{fig:ablation_both}
\end{figure}

To further evaluate our approach, we conduct extensive ablation studies. All experiments are performed with a fixed number of seeds to ensure reproducibility by eliminating the effect of random initialisation.

\paragraph{\textbf{The range of human arms in training}}

The most important part of our architecture is the pseudo-depth-based distance segmentation, which aims to remove irrelevant information from the processed scene, except for the human hands and the manipulated object. It raises the key question of what value of distance should be used as the threshold $t$. In the case of pseudo depth obtained with \textit{DPT-Hybrid}, the depth values are normalised, where $t \in <0,1>$. To select $t$, we first observe the dataset samples and choose values that lead to the preservation of hands and objects only. However, as it is based on estimation, the behaviour is not the same for all samples for the same $t$ and none of these values can be considered good without being proven with performance. In the second step, we search for the best performance by retraining the architecture for each of these $t \in \{0.35,0.39,0.43,0.47,0.51\}$. The results highlight $t=0.47$ as the highest performance value and we observe the performance decrease above and below this value, proving the usability of the proposed method. All results are presented in the left sub-figure of Fig. \ref{fig:ablation_both}.

\paragraph{\textbf{The range of human arms in inference}}

Following the choice of $t$ in training, we examine the choice of t in testing for the best-performing model with $t=0.47$ in training. We run tests for $t \in \{0.35,0.39,0.43,0.47,0.51\}$. All results are shown in the right subplot of Fig. \ref{fig:ablation_both}. The effect of $t$ is significantly lower than in training and does not affect performance much.

\paragraph{\textbf{Pseudo-depth-based SHARP module}}

\begin{figure}[t]
    \centering
    
    \subfigure[]{
        \includegraphics[height=2.6cm]{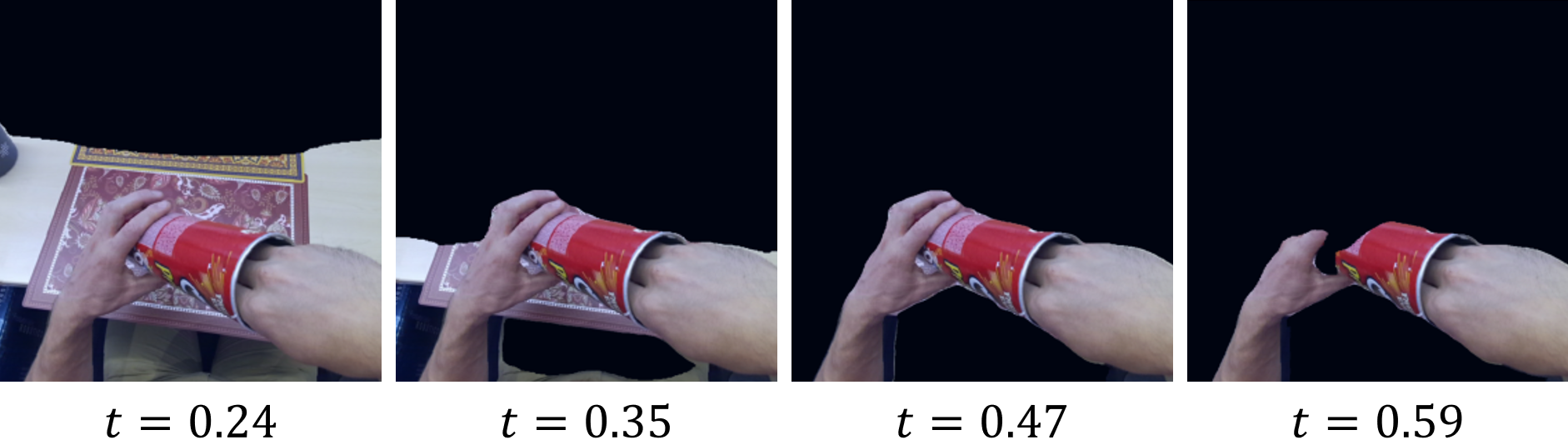}
        \label{fig:different_tresholds_qual:subplot1}
    }
    \hfill
    \subfigure[]{
        \includegraphics[height=2.6cm]{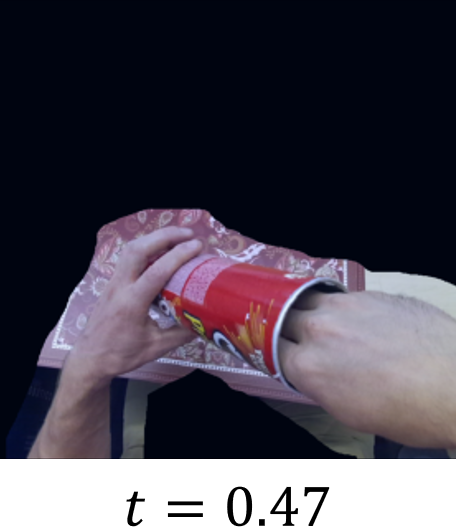}
        \label{fig:different_tresholds_qual:subplot2}
    }
    \caption{\
    On the left, frame processed with \textit{SHARP} and different values of $t$. On the right, the same frame processed with \textit{SHARP}, $t = 0.47$ and with de-sharpening applied.
    }
    \label{fig:different_tresholds_qual}
\end{figure}

\begin{table}[t]
\centering

\caption{Results of ablations studies with different depth image types used in \textit{SHARP}. All results provided in \textit{mm} in camera space for left, right and both hands.}
\label{tab:ablation}

\begin{tabular*}{\textwidth}{l@{
\extracolsep{\fill}}cccc}

\toprule
       & Depth    & MPJPE Left $\downarrow$& MPJPE Right $\downarrow$& MPJPE Both $\downarrow$\\
\midrule
Ours & Estimated & 30.31     & 27.02  & 28.66    \\ 

Ablation I & \ding{55} & 32.95     & 38.01  & 35.48    \\ 

Ablation II & Ground Truth & 21.31     & 28.86  & 25.09    \\ 

Ablation III &Est.+De-sharpen    & 39.49       & 35.01 &   37.25    \\ 
\bottomrule

\end{tabular*}%
\end{table}
 
We evaluate \textit{SHARP}'s impact on the egocentric 3D hand pose estimation performance. The proposed architecture is retrained according to the previously described process without the \textit{SHARP} module, using only unsegmented RGB images representing the full scene. The network reduced by the \textit{SHARP} module in a fixed seed run achieves an MPJPE of 35.48 mm compared to 28.66 mm obtained with \textit{SHARP}. The result is referenced in Table \ref{tab:ablation} as \textit{Ablation I}. The process is repeated three times to reduce the random effect of network initialisation and to strengthen the justification of the idea. The average of the three runs without the \textit{SHARP} module is $35.34 \pm 0.17$, while with \textit{SHARP}, the performance improves to $29.61\pm0.71$ mm, demonstrating the high importance of the proposed architecture.

\paragraph{\textbf{Oracle depth-based SHARP module}}

\textit{SHARP} uses the state-of-the-art depth estimation network \textit{DPT-Hybrid}. Like any deep learning architecture, this model is prone to errors. On the other hand, with progress in architecture development, depth estimation networks will improve in the future, leading to an improvement in the performance of our method. To highlight this potential, we retrain the network with an oracle ground truth depth image provided in the \textit{H2O Dataset}. The depth image represents the distance in mm from a camera. For this reason, we choose $t=700$ mm. The results are superior, achieving an MPJPE of 25.09 mm, better than any state-of-the-art method at the time of this study. The experiment is referred as \textit{Ablation II} in Table \ref{tab:ablation}. This performance demonstrates the potential of our approach when fed with less noisy pseudo-depth data.

\paragraph{\textbf{De-sharpening of segmentation mask}}

The segmentation mask, derived from a pseudo-depth scene representation, consists of sharp edges surrounding the human arms and the manipulated object, based on a distance. Depth estimation is prone to error, and in some scenes, this sharp-edge segmentation leads to the loss of parts of the image that represent relevant information, e.g. human hand. This negative effect can be reduced in two ways, by changing the segmentation threshold as shown in Fig. \ref{fig:different_tresholds_qual:subplot1} or by de-sharpening the edges. The effect of the de-sharpening process is presented in Fig. \ref{fig:different_tresholds_qual:subplot2}. In this ablation, we observe the effect of edge de-sharpening by blurring the mask derived from the pseudo-depth scene representation. Performance drops to 37.25 mm, highlighting the usefulness of the \textit{SHARP} module only with accurate masking.

\section{Conclusion}
\label{sec:conclusion}
In this study, a 3D hand pose estimation model has been developed for the egocentric perspective. The novelty of the proposed architecture lies in the \textit{SHARP} module, which uses pseudo-depth scene representation obtained through a monocular depth estimation model. Thanks to the characteristic of a fixed camera to a user in the egocentric perspective and a constant range of human arms, the distance information is used to remove irrelevant information from the scene. Experiments with our network showed an improvement in performance of 7 mm in the MPJPE metric when using \textit{SHARP}, with the best result of MPJPE equal to 28.66 mm placing as the second best result on the \textit{H2O Dataset}. The further potential of the \textit{SHARP} module was confirmed with the use of the ground truth depth image, resulting in the best result of all state-of-the-art methods equal to 25.09 mm. Furthermore, estimated 3D hand poses were used alongside object detection as input for the action recognition model, where each frame is described by a vector containing the 3D hand pose and the object bounding box, and their sequence is embedded using a transformer-based network. The results obtained on \textit{H2O Dataset}, which includes actions where one hand or two hands interact with objects, resulted in 91.73\% accuracy, outperforming the state-of-the-art.

Our study shows that using pseudo depth to remove irrelevant information in the egocentric scene with current state-of-the-art monocular depth estimation methods improves 3D hand pose performance. The quality of pseudo depth correlates with pose estimation error and requires a sharp and accurate representation of human hands in the scene. In the future, with the advancement of depth estimation networks, this approach has a chance to improve hand pose estimation tasks further, leading to more accurate action recognition.

\subsubsection{Acknowledgements} 

Part of this work was conducted during Wiktor's research secondment at the University of Bristol within the Machine Learning and Computer Vision Research Group (MaVi). We thank the group for their support and resources. This research was supported by VisuAAL ITN H2020 (grant agreement no. 861091) and the Austrian Research Promotion Agency (grant agreement no. 49450173).

\bibliographystyle{splncs04}
\bibliography{bib_file}

\begin{thebibliography}{10}
\providecommand{\url}[1]{\texttt{#1}}
\providecommand{\urlprefix}{URL }
\providecommand{\doi}[1]{https://doi.org/#1}

\bibitem{aboukhadra2023thor-net}
Aboukhadra, A., Malik, J., Elhayek, A., Robertini, N., Stricker, D.: {THOR-Net: End-to-end Graformer-based Realistic Two Hands and Object Reconstruction with Self-supervision}. In: Proceedings of the IEEE/CVF Winter Conference on Applications of Computer Vision. pp. 1001--1010 (01 2023). \doi{10.1109/WACV56688.2023.00106}

\bibitem{carreira2017quo}
Carreira, J., Zisserman, A.: {Quo Vadis, Action Recognition? a New Model and the Kinetics Dataset}. In: Proceedings of the IEEE Conference on Computer Vision and Pattern Recognition. pp. 6299--6308 (2017). \doi{10.1109/CVPR.2017.502}

\bibitem{cartas2017contextually}
Cartas, A., Radeva, P., Dimiccoli, M.: {Contextually Driven First-Person Action Recognition from Videos}. In: Presentation at EPIC@ ICCV2017 Workshop. p.~8 (2017)

\bibitem{chen2016single}
Chen, W., Fu, Z., Yang, D., Deng, J.: {Single-Image Depth Perception in the Wild}. Advances in Neural Information Processing Systems  \textbf{29} (2016)

\bibitem{cho2023transformer}
Cho, H., Kim, C., Kim, J., Lee, S., Ismayilzada, E., Baek, S.: {Transformer-Based Unified Recognition of Two Hands Manipulating Objects}. In: Proceedings of the IEEE/CVF Conference on Computer Vision and Pattern Recognition. pp. 4769--4778 (2023). \doi{10.1109/CVPR52729.2023.00462}

\bibitem{Damen2018EPICKITCHENS}
Damen, D., Doughty, H., Farinella, G.M., Fidler, S., Furnari, A., Kazakos, E., Moltisanti, D., Munro, J., Perrett, T., Price, W., Wray, M.: {Scaling Egocentric Vision: The EPIC-KITCHENS Dataset}. In: European Conference on Computer Vision (ECCV) (2018). \doi{10.1007/978-3-030-01225-0_44}

\bibitem{das2021symmetric}
Das, P., Ortega, A.: {Symmetric Sub-graph Spatio-temporal Graph Convolution and its Application in Complex Activity Recognition}. In: ICASSP 2021-2021 IEEE International Conference on Acoustics, Speech and Signal Processing (ICASSP). pp. 3215--3219. IEEE (2021). \doi{10.1109/ICASSP39728.2021.9413833}

\bibitem{dosovitskiy2020image}
Dosovitskiy, A., Beyer, L., Kolesnikov, A., Weissenborn, D., Zhai, X., Unterthiner, T., Dehghani, M., Minderer, M., Heigold, G., Gelly, S., Uszkoreit, J., Houlsby, N.: {An Image is Worth 16x16 Words: Transformers for Image Recognition at Scale}. In: International Conference on Learning Representations (2021)

\bibitem{eigen2014depth}
Eigen, D., Puhrsch, C., Fergus, R.: {Depth Map Prediction from a Single Image using a Multi-Scale Deep Network}. Advances in Neural Information Processing Systems  \textbf{27} (2014)

\bibitem{feichtenhofer2019slowfast}
Feichtenhofer, C., Fan, H., Malik, J., He, K.: {Slowfast Networks for Video Recognition}. In: Proceedings of the IEEE/CVF International Conference on Computer Vision. pp. 6202--6211 (2019). \doi{10.1109/ICCV.2019.00630}

\bibitem{garcia2018first}
Garcia-Hernando, G., Yuan, S., Baek, S., Kim, T.K.: {First-Person Hand Action Benchmark With RGB-D Videos and 3D Hand Pose Annotations}. In: Proceedings of the IEEE Conference on Computer Vision and Pattern Recognition. pp. 409--419 (2018). \doi{10.1109/CVPR.2018.00050}

\bibitem{garg2016unsupervised}
Garg, R., Bg, V.K., Carneiro, G., Reid, I.: {Unsupervised CNN for Single View Depth Estimation: Geometry to the Rescue}. In: Computer Vision--ECCV 2016: 14th European Conference, Amsterdam, The Netherlands, October 11-14, 2016, Proceedings, Part VIII 14. pp. 740--756. Springer (2016). \doi{10.1007/978-3-319-46484-8_45}

\bibitem{godard2019digging}
Godard, C., Mac~Aodha, O., Firman, M., Brostow, G.J.: {Digging into self-supervised monocular depth estimation}. In: Proceedings of the IEEE/CVF International Conference on Computer Vision. pp. 3828--3838 (2019). \doi{10.1109/ICCV.2019.00393}

\bibitem{grauman2022ego4d}
Grauman, K., Westbury, A., Byrne, E., Chavis, Z., Furnari, A., Girdhar, R., Hamburger, J., Jiang, H., Liu, M., Liu, X., et~al.: {Ego4D: Around the World in 3,000 Hours of Egocentric Video}. In: Proceedings of the IEEE/CVF Conference on Computer Vision and Pattern Recognition. pp. 18995--19012 (2022). \doi{10.1109/CVPR52688.2022.01842}

\bibitem{hasson2020leveraging}
Hasson, Y., Tekin, B., Bogo, F., Laptev, I., Pollefeys, M., Schmid, C.: {Leveraging Photometric Consistency over Time for Sparsely Supervised Hand-Object Reconstruction}. In: Proceedings of the IEEE/CVF Conference on Computer Vision and Pattern Recognition. pp. 571--580 (2020). \doi{10.1109/CVPR42600.2020.00065}

\bibitem{Kwon_2021_ICCV}
Kwon, T., Tekin, B., St\"uhmer, J., Bogo, F., Pollefeys, M.: {H2O: Two Hands Manipulating Objects for First Person Interaction Recognition}. In: Proceedings of the IEEE/CVF International Conference on Computer Vision (ICCV). pp. 10138--10148 (October 2021). \doi{10.1109/iccv48922.2021.00998}

\bibitem{text2taste}
Mucha, W., Cuconasu, F., Etori, N.A., Kalokyri, V., Trappolini, G.: {TEXT2TASTE: A Versatile Egocentric Vision System for Intelligent Reading Assistance Using Large Language Model}. In: Computers Helping People with Special Needs. pp. 285--291. Springer Nature Switzerland, Cham (2024). \doi{10.1007/978-3-031-62849-8_35}

\bibitem{effhandegonet}
Mucha, W., Kampel, M.: {In My Perspective, in My Hands: Accurate Egocentric 2D Hand Pose and Action Recognition}. In: 2024 IEEE 18th International Conference on Automatic Face and Gesture Recognition (FG). pp.~1--9 (2024). \doi{10.1109/FG59268.2024.10582035}

\bibitem{mueller2017real}
Mueller, F., Mehta, D., Sotnychenko, O., Sridhar, S., Casas, D., Theobalt, C.: {Real-time Hand Tracking Under Occlusion From an Egocentric RGB-D Sensor}. In: Proceedings of the IEEE International Conference on Computer Vision. pp. 1154--1163 (2017). \doi{10.1109/CVPR.2019.01231}

\bibitem{nguyen2019neural}
Nguyen, X.S., Brun, L., L{\'e}zoray, O., Bougleux, S.: {A Neural Network Based on SPD Manifold Learning for Skeleton-based Hand Gesture Recognition}. In: Proceedings of the IEEE/CVF Conference on Computer Vision and Pattern Recognition. pp. 12036--12045 (2019). \doi{10.1109/CVPR.2019.01231}

\bibitem{nunez2022egocentric}
N{\'u}{\~n}ez-Marcos, A., Azkune, G., Arganda-Carreras, I.: {Egocentric Vision-based Action Recognition: a Survey}. Neurocomputing  \textbf{472},  175--197 (2022). \doi{10.1016/j.neucom.2021.11.081}

\bibitem{ohkawa:cvpr23}
Ohkawa, T., He, K., Sener, F., Hodan, T., Tran, L., Keskin, C.: {AssemblyHands:} towards egocentric activity understanding via 3d hand pose estimation. In: Proceedings of the IEEE/CVF Conference on Computer Vision and Pattern Recognition (CVPR). pp. 12999--13008 (2023). \doi{10.1109/CVPR52729.2023.01249}

\bibitem{ranftl2021vision}
Ranftl, R., Bochkovskiy, A., Koltun, V.: {Vision Transformers for Dense Prediction}. In: {Proceedings of the IEEE/CVF International Conference on Computer Vision}. pp. 12179--12188 (2021). \doi{10.1109/ICCV48922.2021.01196}

\bibitem{ranftl2020towards}
Ranftl, R., Lasinger, K., Hafner, D., Schindler, K., Koltun, V.: {Towards Robust Monocular Depth Estimation: Mixing Datasets for Zero-shot Cross-dataset Transfer}. IEEE Transactions on Pattern Analysis and Machine Intelligence  \textbf{44}(3),  1623--1637 (2020). \doi{10.1109/TPAMI.2020.3019967}

\bibitem{tan2021efficientnetv2}
Tan, M., Le, Q.: {Efficientnetv2: Smaller Models and Faster Training}. In: International Conference on Machine Learning. pp. 10096--10106. PMLR (2021)

\bibitem{tekin2019h+}
Tekin, B., Bogo, F., Pollefeys, M.: {H+O: Unified Egocentric Recognition of 3D Hand-object Poses and Interactions}. In: Proceedings of the IEEE/CVF Conference on Computer Vision and Pattern Recognition. pp. 4511--4520 (2019). \doi{10.1109/CVPR.2019.00464}

\bibitem{wang2022yolov7}
Wang, C.Y., Bochkovskiy, A., Liao, H.Y.M.: {YOLOv7: Trainable Bag-of-Freebies Sets New State-of-the-Art for Real-Time Object Detectors}. In: Proceedings of the IEEE/CVF Conference on Computer Vision and Pattern Recognition. pp. 7464--7475 (2023). \doi{10.48550/arXiv.2207.02696}

\bibitem{wang2018non}
Wang, X., Girshick, R., Gupta, A., He, K.: {Non-local Neural Networks}. In: Proceedings of the IEEE Conference on Computer Vision and Pattern Recognition. pp. 7794--7803 (2018). \doi{10.1109/CVPR.2018.00813}

\bibitem{HoloAssist2023}
Wang, X., Kwon, T., Rad, M., Pan, B., Chakraborty, I., Andrist, S., Bohus, D., Feniello, A., Tekin, B., Frujeri, F.V., Joshi, N., Pollefeys, M.: {HoloAssist: an Egocentric Human Interaction Dataset for Interactive AI Assistants in the Real World}. In: Proceedings of the IEEE/CVF International Conference on Computer Vision (ICCV). pp. 20270--20281 (October 2023). \doi{10.1109/ICCV51070.2023.01854}

\bibitem{wen2023hierarchical}
Wen, Y., Pan, H., Yang, L., Pan, J., Komura, T., Wang, W.: {Hierarchical Temporal Transformer for 3D Hand Pose Estimation and Action Recognition from Egocentric RGB Videos}. In: Proceedings of the IEEE/CVF Conference on Computer Vision and Pattern Recognition. pp. 21243--21253 (2023). \doi{10.1109/CVPR52729.2023.02035}

\bibitem{yamazaki2017hand}
Yamazaki, W., Ding, M., Takamatsu, J., Ogasawara, T.: {Hand Pose Estimation and Motion Recognition Using Egocentric RGB-D Video}. In: 2017 IEEE International Conference on Robotics and Biomimetics (ROBIO). pp. 147--152. IEEE (2017). \doi{10.1109/ROBIO.2017.8324409}

\bibitem{yan2018spatial}
Yan, S., Xiong, Y., Lin, D.: {Spatial Temporal Graph Convolutional Networks for Skeleton-based Action Recognition}. In: Proceedings of the AAAI Conference on Artificial Intelligence. vol.~32 (2018). \doi{10.1609/aaai.v32i1.12328}

\end{thebibliography}

\end{document}